\documentclass{article}
\usepackage{spconf, amsmath, graphicx}
\usepackage{amssymb}
\usepackage{enumitem}
\usepackage{xcolor}
\usepackage{subcaption}

\newcommand{\bl}{\boldsymbol{l}}

\setlist{nosep,leftmargin=14pt}

\title{DiffuSAM: Diffusion-Based Prompt-Free SAM2 for\\Few-Shot and Source-Free Medical Image Segmentation}

\name{Tal Grossman$^{1, \dagger}$\thanks{Corresponding author (e-mail): \texttt{talgrossman22@gmail.com}.}, Noa Cahan$^{1, \dagger}$, Lev Ayzenberg$^{1}$, Hayit Greenspan$^{2}$}
\address{$^{1}$School of Electrical Engineering, Tel Aviv University, Israel\\
$^{2}$School of Biomedical Engineering, Tel Aviv University, Israel}

\begin{document}

  \maketitle  

\begingroup
\renewcommand\thefootnote{}\footnote{$^\dagger$ These authors contributed equally to this work.}
\addtocounter{footnote}{-1}
\endgroup

    \begin{abstract}
Segmentation models such as Segment Anything Model (SAM) and SAM2 achieve strong prompt-driven zero-shot performance. However, their training on natural images limits domain transfer to medical data. Consequently, accurate segmentation typically requires extensive fine-tuning and expert-designed prompts. We propose DiffuSAM, a diffusion-based adaptation of SAM2 for prompt-free medical image segmentation. Our framework synthesizes SAM2-compatible segmentation mask-like embeddings via a lightweight diffusion-prior from off-the-shelf frozen SAM2 image features. The generated embeddings are integrated into SAM2’s mask decoder to produce accurate segmentations, thereby eliminating the need for user prompts. The diffusion prior is further conditioned on previously segmented slices, enforcing spatial consistency across volumes. Evaluated on the BTCV and CHAOS datasets for CT and MRI under Source-Free Unsupervised Domain Adaptation (SF-UDA) and Few-Shot settings, DiffuSAM achieves competitive performance with efficient training and inference. Code is available upon request from the corresponding author.
  \end{abstract}


\section{Introduction}
Medical image segmentation is essential for clinical diagnosis and treatment planning. Deep learning has achieved remarkable success in this task \cite{RFB15a}, but supervised approaches require large annotated datasets, which are costly and expert-dependent. Foundation models such as SAM \cite{kirillov2023segany} and SAM2 \cite{ravi2024sam2} exhibit strong zero-shot segmentation on natural images but face challenges when applied to medical data due to modality differences in texture, contrast, and anatomy. In addition, their reliance on user prompts limits practical use in clinical workflows. Previous studies have adapted SAM or SAM2 to medical data \cite{MedSAM}, yet fine-tuning is computationally demanding and often still requires prompts. Recent prompt-free variants \cite{ayzenberg2024protosamoneshotmedicalimage} were developed for SAM, but to our knowledge, none target SAM2.

Diffusion models have recently emerged as powerful generative frameworks capable of modeling complex distributions across imaging modalities \cite{DING2024102540}. Latent diffusion in particular accelerates training and sampling by operating in lower-dimensional feature spaces.  

To address these limitations and harness generative models advantages, we propose a conditional latent diffusion framework that builds on SAM2’s strong zero-shot generalization for prompt-free medical image segmentation. This ability aligns with the goal of unsupervised domain adaptation (UDA), and particularly with SF-UDA, which seeks to adapt pre-trained models to new domains without requiring source data \cite{DFG}. We train a UNet-based diffusion model to generate SAM2 memory embeddings conditioned on the frozen SAM2 image encoder output. The synthesized memory is injected into the SAM2 memory attention and mask decoder to produce segmentation maps without prompts. The framework, termed \textbf{DiffuSAM}, learns structured memory representations that capture anatomical variability and enable consistent, prompt-free segmentation across domains.

Our main contributions are summarized below:
\begin{itemize}
  \item A novel diffusion-based framework for prompt-free medical image segmentation in SAM2.  
  \item We operate in the latent embedding space of SAM2, allowing the model to capture high-level feature distributions.
 \item Integration with SAM2’s volumetric memory attention ensures anatomical continuity across slices.  
 \item Demonstrated performance on BTCV and CHAOS datasets for CT and MRI segmentation under SF-UDA and few-shot settings.  
\end{itemize}

  \section{Method}

\begin{figure}[t]
  \centering
  \includegraphics[width=0.45\textwidth]{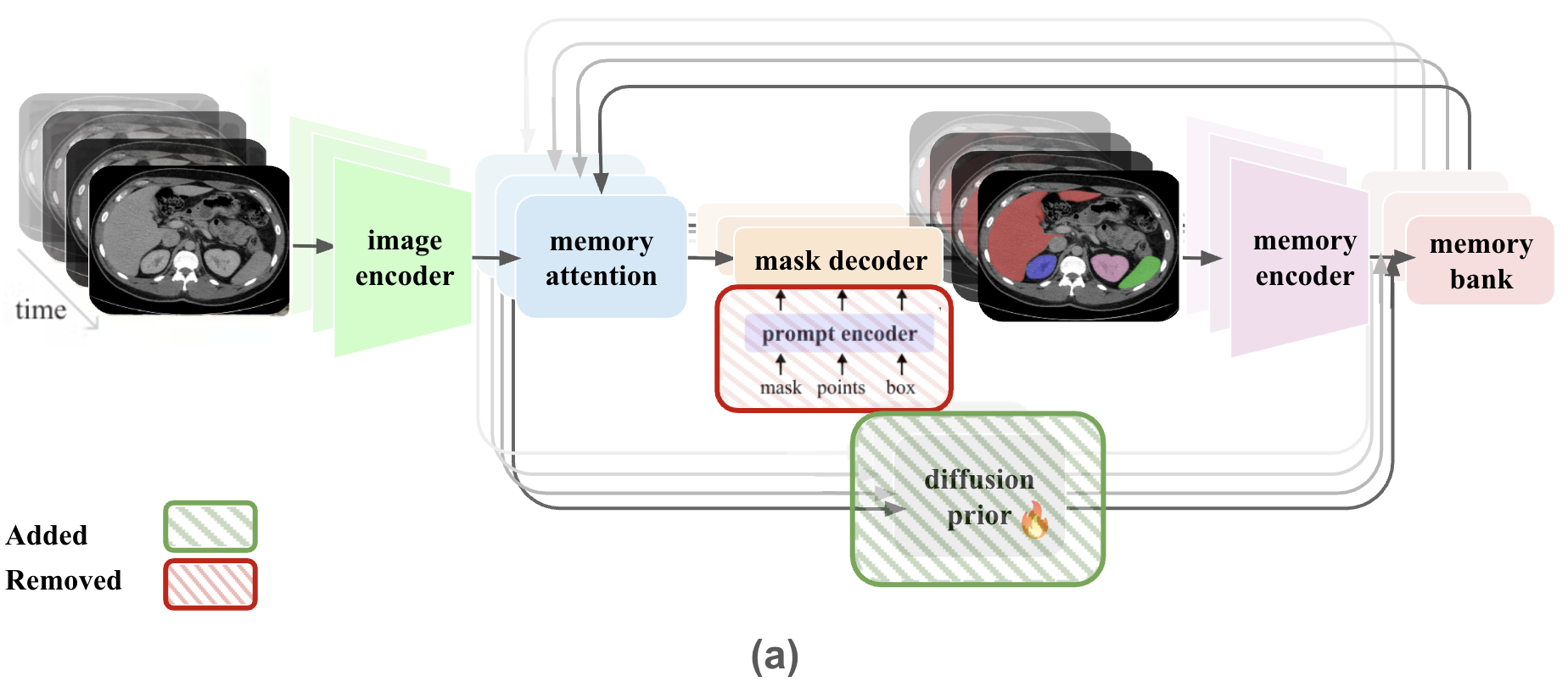}
    \includegraphics[width=0.45\textwidth]{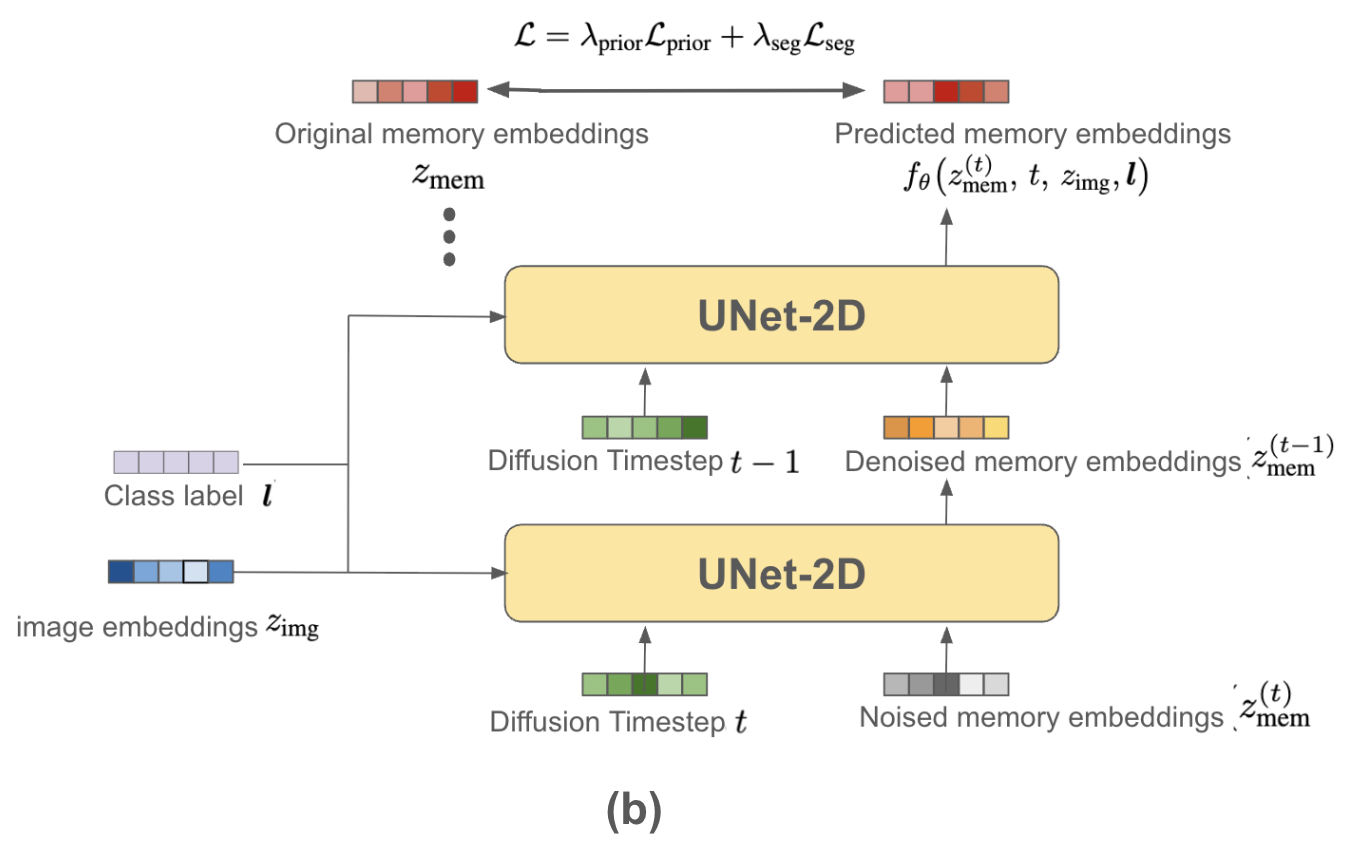}
\caption{An outline of our proposed method. (a) In our proposed network, we add a diffusion prior (in Green) to generate memory embeddings from the image embeddings, thereby removing the need for a user prompt (in Red). During training, the SAM2 components all remain frozen. (b) Diffusion Prior model. During training, a diffusion model takes a memory embedding to which noise has been added, the matching image embedding, an embedding of the current time step, and a label. The system learns to use the image embedding to remove the noise in successive time steps. In inference, it generates a memory embedding by starting with pure noise and an image embedding and removing noise iteratively according to that embedding.}
  \label{fig:pipeline}
\end{figure}

  An overview of DiffuSAM is shown in Fig. \ref{fig:pipeline}. (a). Our model was inspired by \cite{CahanSizikov}, in which a diffusion-prior was trained for cross-modal embedding generation for classification enhancement. During training, we fit a lightweight diffusion model to generate the latent SAM2 memory embeddings, see Fig.~\ref{fig:pipeline}. (b). At inference, the learned prior generates memory embeddings conditioned on the SAM2 image encoder embeddings. These synthesized embeddings are then injected into the pre-trained SAM2 memory attention and mask decoder to produce segmentation masks without any user prompts.

  \subsection{Diffusion-based SAM2 Memory Embeddings Generation}

  Each training slice is passed through the pre-trained SAM2 image encoder and the SAM2 memory encoder ($\mathrm{ME}$) to obtain the image embedding $z_{\mathrm{img}}$ and the memory embedding $z_{\mathrm{mem}}$, respectively. Let $\bl$ denote the class label for the slice. We learn a conditional prior
  $p_\theta\big(z_{\mathrm{mem}}\mid z_{\mathrm{img}}, \bl\big)$
  that generates memory embeddings given the image embedding and label. Concretely, we model the continuous vector $z_{\mathrm{mem}}$ with a Gaussian diffusion model conditioned on slice embedding and class label.
  To achieve this, we train a UNet2D diffusion prior under a Denoising Diffusion Probabilistic Model (DDPM) noise schedule on a sequence comprising the image slice, an embedding of the diffusion timestep, the noised memory embedding, class label embeddings, and a final embedding whose output is used to predict the denoised memory embedding, then use the scheduler to sample predicted $\hat{z_{\mathrm{mem}}}$. 

  The block diagram for the diffusion prior is depicted in  Fig.~\ref{fig:pipeline}.(b). 
  \begin{equation}
    \mathcal{L}_{\text{prior}}
    = \mathbb{E}_{t\sim[1,T],\, z_{\text{mem}}^{(t)}\sim q_t}
      \big\| f_{\theta}\big(z_{\text{mem}}^{(t)},\, t,\, z_{\text{img}}, \bl\big)
      - z_{\text{mem}} \big\|_2^2
  \end{equation}
  In addition, we incorporate a segmentation loss $\mathcal{L}_{\text{seg}}$ computed on the decoded masks obtained by passing the synthesized memory embeddings $\hat{z}_{\mathrm{mem}}$ through the pre-trained SAM2 memory attention $\mathrm{MA}$ and mask decoder $\mathrm{D}$ along with the image embeddings $z_{\mathrm{img}}$. The overall training objective is a weighted combination of the prior loss and segmentation loss:
  \begin{equation}
    \mathcal{L} = \lambda_{\text{prior}} \mathcal{L}_{\text{prior}} +
    \lambda_{\text{seg}} \mathcal{L}_{\text{seg}}
  \end{equation}

  \subsection{Volumetric Consistency via Cross-Slice Conditioning}
  To enforce anatomical consistency and spatial coherence across adjacent slices in 3D medical volumes, we extend the diffusion prior to condition on memory embeddings from neighboring slices. Specifically, when generating the memory embedding for a given slice $i$, we encode the final SAM2-decoded segmentation predictions from adjacent slices $adj$ using the memory encoder $\mathrm{ME}$ to produce volumetric memory embeddings. These embeddings are then fused with the slice-specific memory to create unified embeddings that incorporate both anatomical knowledge and volumetric consistency. The starting slice is the middle slice in the axial direction, and we propagate forward and backward to the volume edges. Finally, the predicted mask for slice $i$ is obtained as:
  \begin{equation}
    \label{sam2_volume_consistency_eq}
    \hat{y_i} = \mathrm{D}\left\{ \mathrm{MA}\left[ \hat{z}_{\mathrm{mem},i}, \mathrm{ME}\left( \hat{y_{adj}}, z_{\mathrm{img},\mathrm{adj}} \right) \right], z_{\mathrm{img},i} \right\}
  \end{equation}
  where $\hat{y_i}$ is the decoded mask for slice $i$, $\hat{z}_{\mathrm{mem}, i}$ is the synthesized memory embedding for slice $i$, and $\hat{y_{adj}}$ and $z_{\mathrm{img},\mathrm{adj}}$ are the decoded masks and image embeddings for adjacent slices, respectively. 

  As for the Source-Free unsupervised-domain-adaptation setting, we leverage the feature affinity property observed in prior works \cite{DFG}. Specifically, the target domain features of the same class produced by the source model tend to be closely located despite the domain gap, which we refer to as the feature affinity property. SAM2 image embeddings exhibit this property, allowing us to directly apply the source-trained diffusion prior on target domain slices without further fine-tuning.
  We first generate pseudo-memory embeddings for all target domain slices, then we process each target slice in the same manner as in Eq.~\ref{sam2_volume_consistency_eq}, using the synthesized memory embeddings and the SAM2 image embeddings from the target domain. 

  \section{Experiments}
  \subsection{Dataset and setup}

  We evaluate our method on abdominal imaging datasets. The Beyond the Cranial Vault (BTCV) dataset \cite{BTCV} consists of 30 CT volumes, while the CHAOS dataset \cite{CHAOS2021} comprises 20 T2-SPIR MRI volumes. Both datasets provide multi-organ annotations, specifically, we focus on segmenting the spleen, right kidney, left kidney, and liver.
  Volumes are resampled to 2D axial slices $512{\times}512$ to fit the SAM2 tiny configuration.
  We report Dice on the test set for both few-shot and SF-UDA settings.
  For few-shot experiments, we randomly split the BTCV data into 10\% training and 90\% testing, resulting in 3 volumes for training and 27 for testing.
  In the SF-UDA setting, we train on the BTCV CT data and test on the CHAOS MRI data without access to source domain images.

  The implementation is based on PyTorch and runs on an NVIDIA RTX A5000 GPU, requiring less than 5 GB of GPU memory for both training and inference. The framework trains for 7800 iterations with a batch size of 4, using the Adam optimizer with a learning rate of $1 \times 10^{-4}$. The noise schedule follows a constant schedule with 1000 diffusion steps. The weights for the loss terms are set to $\lambda_{\text{prior}} = 1$ and $\lambda_{\text{seg}} = 1$. During inference, we use only 2 diffusion steps for memory generation. We used SAM2 tiny model as the backbone for all experiments.

  \subsection{Results}
\begin{figure}[t]
  \centering
  \begin{minipage}[b]{0.495\linewidth}
    \centering
    \includegraphics[width=\linewidth]{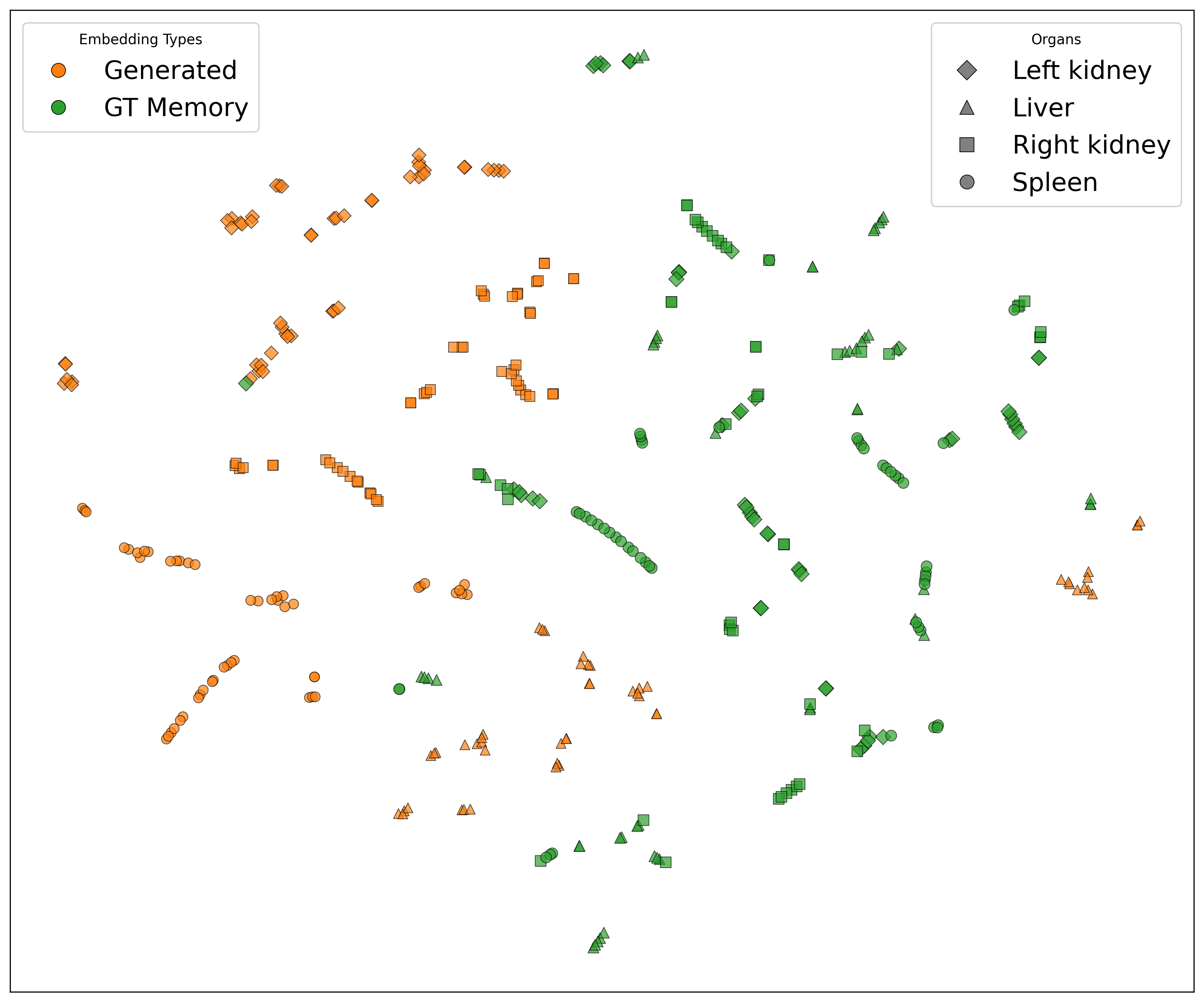}
    \\[-2pt]{\footnotesize (A)}
  \end{minipage}\hfill
  \begin{minipage}[b]{0.495\linewidth}
    \centering
    \includegraphics[width=\linewidth]{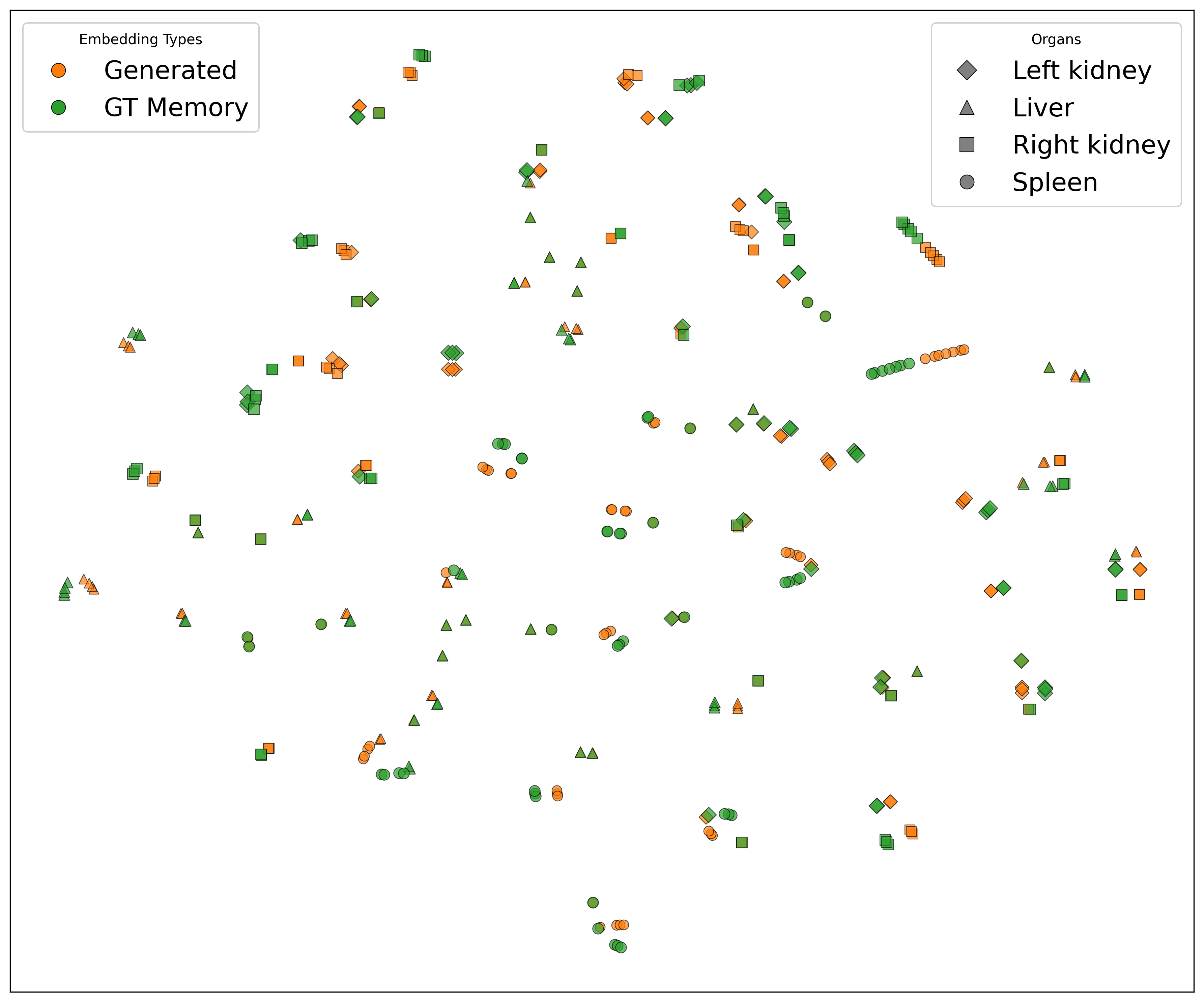}
    \\[-2pt]{\footnotesize (B)}
  \end{minipage}
  \caption{t-SNE visualizations of generated vs. ground-truth SAM2 memory embeddings at the beginning (A) and end (B) of training.}
  \label{fig:tsne-evolution}
\end{figure}

  In Figure~\ref{fig:tsne-evolution}, we show 2D t-SNE of SAM2 memory embeddings: diffusion-generated vs. ground truth on held-out slices. At early training stage (A), the generated embeddings from diffuse, partially misaligned clusters; by the end of the training stage (B), clusters are compact and nearly co-located with ground truth, indicating a learned class-conditioned mapping into SAM2’s memory space.

  \begin{figure}[t]
  \centering
  \includegraphics[width=0.95\linewidth]{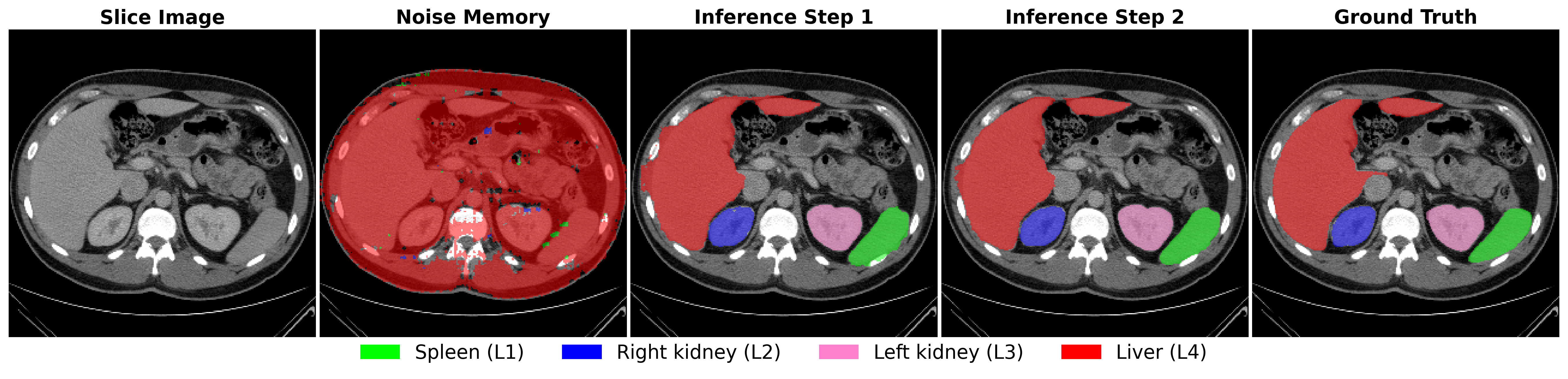}\\[4pt]
  \includegraphics[width=0.95\linewidth]{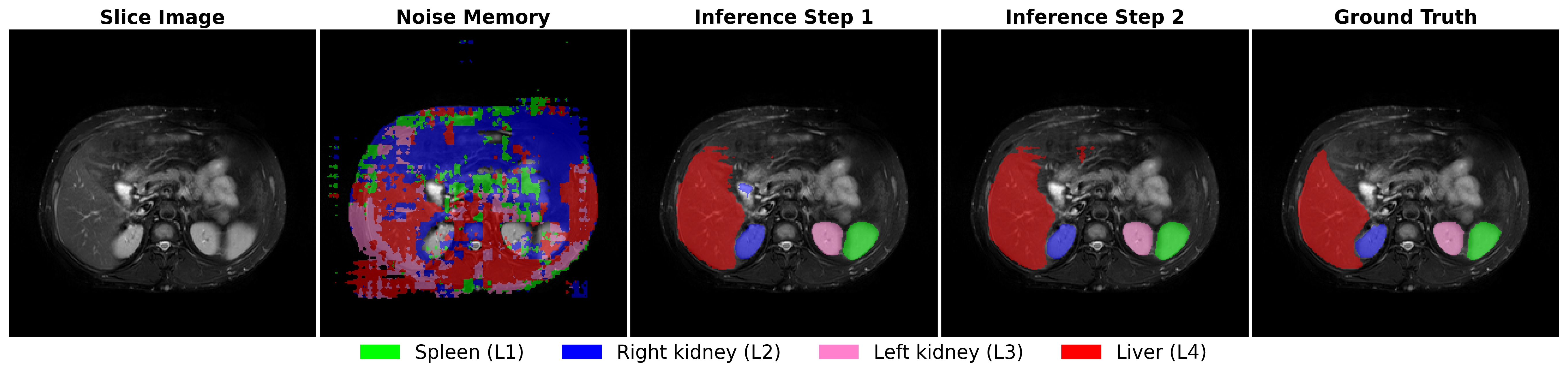}
  \caption{Diffusion inference steps. Top: CT; Bottom: MRI.}
  \label{fig:inference-steps}
  \end{figure}

  For the few-shot setting, we compared DiffuSAM with supervised methods, including UNet \cite{RFB15a} and UNETR \cite{hatamizadeh2022unetr}, as well as native SAM-based methods, such as SAM \cite{kirillov2023segany}, SAM2 \cite{ravi2024sam2}, and FATE-SAM \cite{he2025fewshotadaptationtrainingfreefoundation}. For supervised methods, we trained the models on the same few-shot training split as DiffuSAM. 
  SAM-based models were evaluated using the frozen pre-trained weights.
  Since these methods rely on user prompts, we used random points from the ground-truth masks as input prompts. 
  FATE-SAM \cite{he2025fewshotadaptationtrainingfreefoundation} conditions inference on similar ground-truth slices, achieving strong results but depending on training data at test time.
  Table \ref{tab:fewshot}.a. summarizes the differences in model data setting and Table \ref{tab:fewshot}.b. presents the Dice results per organ on the BTCV dataset in the few-shot setting, with results for baseline methods adopted from FATE-SAM \cite{he2025fewshotadaptationtrainingfreefoundation}.
  Our method outperforms all other methods, achieving an average Dice of 87.24\%, demonstrating the effectiveness of our diffusion-based memory generation approach for prompt-free medical image segmentation.
  In addition, we performed an ablation experiment to assess the impact of volumetric consistency conditioning on adjacent slices. Results indicate that incorporating cross-slice conditioning improves segmentation accuracy. DiffuSAM-no 3D refer to the model with diffusion prior only, i.e., without volumetric conditioning. It shows that adding volumetric consistency boosts the average Dice from 82.72\% to 87.24\%.
  Diffusion steps during inference are illustrated for CT in Fig.~\ref{fig:inference-steps} top row, showing the progressive refinement of the segmentation mask from noise to the final output.

\begin{table}[h!]
\centering
\caption{Few-shot settings: (a) model requirements and (b) Dice results per organ on BTCV dataset.}
\label{tab:fewshot}
\small

\begin{subtable}[t]{0.48\textwidth}
\centering
\caption{Model data setting}
\resizebox{\columnwidth}{!}{
\begin{tabular}{|c|c|c|c|}
\hline
Method & \shortstack[c]{Natural images  \\ pre-trained} & 
\shortstack[c]{Requires domain \\ specific dataset} & 
\shortstack[c]{Manual prompt} \\
\hline \hline
UNet \cite{RFB15a} & $\times$ & \checkmark & $\times$ \\ \hline
UNETR \cite{hatamizadeh2022unetr} & $\times$ & \checkmark & $\times$ \\ \hline
SAM \cite{kirillov2023segany} & \checkmark & $\times$ & \checkmark \\ \hline
SAM2 \cite{ravi2024sam2} & \checkmark & $\times$ & \checkmark \\ \hline
FATE-SAM \cite{he2025fewshotadaptationtrainingfreefoundation} & \checkmark & \checkmark & $\times$ \\ \hline \hline
Ours (DiffuSAM-no 3D) & \checkmark & \checkmark & $\times$  \\ \hline
Ours (DiffuSAM) & \checkmark & \checkmark & $\times$ \\ \hline
\end{tabular}
}
\end{subtable}
\hfill
\begin{subtable}[t]{0.48\textwidth}
\centering
\caption{Dice scores per organ}
\resizebox{\columnwidth}{!}{
\begin{tabular}{|c|c|c|c|c|c|}
\hline
 & Spleen & R.Kidney & L.Kidney & Liver & \textbf{Avg} \\ \hline \hline
UNet \cite{RFB15a} & 54.85 & 55.99 & 48.70 & 85.85 & 61.35 \\ \hline
UNETR \cite{hatamizadeh2022unetr} & 8.65 & 0.24 & 2.14 & 78.15 & 22.30 \\ \hline
SAM \cite{kirillov2023segany} & 15.53 & 52.19 & 62.19 & 26.56 & 39.12 \\ \hline
SAM2 \cite{ravi2024sam2} & 20.24 & 55.67 & 66.94 & 34.59 & 44.36 \\ \hline
FATE-SAM \cite{he2025fewshotadaptationtrainingfreefoundation} & 86.22 & \textbf{90.99} & \textbf{89.32} & 82.17 & 87.18 \\ \hline \hline
Ours (DiffuSAM-no 3D) & 82.87 & 80.95 & 83.56 & 83.49 & 82.72 \\ \hline
Ours (DiffuSAM) & \textbf{86.54} & 88.61 & 86.50 & \textbf{87.29} & \textbf{87.24} \\ \hline
\end{tabular}
}
\end{subtable}

\end{table}

  In the SF-UDA regime, we compare DiffuSAM to a target-supervised upper bound, and source-free baselines DPL \cite{dpl} and DFG \cite{DFG} (Table \ref{tab:results-SFUDA}). 
  Results for the compared methods are taken as reported in DFG \cite{DFG}.
  DiffuSAM outperforms DPL and performs on par with DFG, a strong feature-affinity-based method, highlighting the competitiveness of our approach. As in the few-shot setting, volumetric cross-slice conditioning improves the average Dice from 80.4\% (DiffuSAM-no 3D) to 85.2\% (DiffuSAM).
  Diffusion steps during inference are illustrated for MRI in Fig.~\ref{fig:inference-steps} bottom row, showing the progressive refinement of the segmentation mask from noise to the final output.

  \begin{table}[h!]
    \caption{SF-UDA setting dice results per organ.}
    \label{tab:results-SFUDA}
    \centering
    \resizebox{\columnwidth}{!}{
    \begin{tabular}{|c|c|c|c|c|c|}
      \hline  
      \multicolumn{6}{|c|}{Source: BTCV; Target: CHAOS (CT $\rightarrow$ MRI)} \\
        \hline \hline
        Method & Spleen & R.Kidney & L.Kidney & Liver & \textbf{Avg} \\
        \hline \hline
        Target-supervised & 94.5 & 95.5 & 95.3 & 95.1 & 95.1 \\
        \hline
        DPL \cite{dpl} & 38.3 & 65.3 & 57.3 & 63.1 & 56.0 \\
        DFG \cite{DFG} & \textbf{85.1} & \textbf{92.5} & 79.6 & \textbf{83.6} & \textbf{85.2} \\
        \hline \hline
        Ours (DiffuSAM-no 3D) & 76.3 & 85.8 & 85.2 & 74.3 & 80.4 \\
        Ours (DiffuSAM) & 84.6 & 87.69 & \textbf{86.29} & 82.2 & \textbf{85.2} \\
      \hline
    \end{tabular}}
  \end{table}

  \section{Discussion and Conclusion}
  We showed that a lightweight diffusion prior over frozen SAM2 features enables prompt-free segmentation with competitive results in few-shot and SF-UDA, with simple training and fast inference. We acknowledge that MedSAM \cite{MedSAM}, trained end-to-end on large medical datasets and optimized for prompts, can outperform ours. This is expected given its domain-specific supervision and prompt guidance, whereas DiffuSAM avoids prompts and full fine-tuning by learning a compact prior in SAM2’s memory space. In the future we plan to explore additional SF-UDA techniques to be integrated with our framework.

  \bibliographystyle{IEEEbib}
  \bibliography{refs}
\end{document}